\documentclass[10pt,twocolumn,letterpaper]{article}

\usepackage[pagenumbers]{cvpr}
\usepackage{algorithm, algpseudocode}

\usepackage{listings}
\usepackage{xcolor}

\definecolor{codegreen}{rgb}{0,0.6,0}
\definecolor{codegray}{rgb}{0.5,0.5,0.5}
\definecolor{codepurple}{rgb}{0.58,0,0.82}
\definecolor{backcolour}{rgb}{0.95,0.95,0.92}

\lstdefinestyle{pytorchstyle}{
    backgroundcolor=\color{backcolour},
    commentstyle=\color{codegreen},
    keywordstyle=\color{magenta},
    numberstyle=\tiny\color{codegray},
    stringstyle=\color{codepurple},
    basicstyle=\ttfamily\footnotesize,
    breakatwhitespace=false,
    breaklines=true,
    captionpos=b,
    keepspaces=true,
    numbers=left,
    numbersep=5pt,
    showspaces=false,
    showstringspaces=false,
    showtabs=false,
    tabsize=2,
    language=Python,
    morekeywords={self, torch, nn, F, with, no_grad}
}

\newcommand{\TODO}[1]{\textbf{\color{red}[TODO: #1]}}

\renewcommand{\TODO}[1]{}

\definecolor{cvprblue}{rgb}{0.21,0.49,0.74}
\usepackage[breaklinks,colorlinks,allcolors=cvprblue]{hyperref}

\def\paperID{}
\def\confName{CVPR}
\def\confYear{2026}

\title{DualTSR: Unified Dual-Diffusion Transformer for\\ Scene Text Image Super-Resolution}
\author{
Axi Niu$^1$ \quad
Kang Zhang$^2$ \quad
Qingsen Yan$^1$ \quad
Hao Jin$^1$ \quad
Jinqiu Sun$^1$ \quad
Yanning Zhang$^1$\\
$^1$Northwestern Polytechnical University\\
$^2$Korea Advanced Institute of Science and Technology\\
{\tt\small \{nax, qingsenyan, sunjinqiu, ynzhang\}@nwpu.edu.cn} \\
{\tt\small jinh@mail.nwpu.edu.cn, zhangkang@kaist.ac.kr} \\
}

\usepackage{amssymb}
\usepackage{pifont}
\usepackage{amsmath,amsfonts,bm}
\usepackage{xcolor}
\usepackage{color, colortbl}
\usepackage{wrapfig}

\usepackage{graphicx}
\usepackage{booktabs}
\usepackage{multirow}

\definecolor{baselinecolor}{gray}{.9}
\newcommand{\cc}[1]{\cellcolor{baselinecolor}{#1}}
\newcommand{\AX}[1]{}
\newcommand{\ZK}[1]{}

\begin{document}
\maketitle
\begin{abstract}
Scene Text Image Super-Resolution (STISR) aims to restore high-resolution details in low-resolution text images, which is crucial for both human readability and machine recognition. Existing methods, however, often depend on external Optical Character Recognition (OCR) models for textual priors or rely on complex multi-component architectures that are difficult to train and reproduce. In this paper, we introduce DualTSR, a unified end-to-end framework that addresses both issues. DualTSR employs a single multimodal transformer backbone trained with a dual diffusion objective. It simultaneously models the continuous distribution of high-resolution images via Conditional Flow Matching and the discrete distribution of textual content via discrete diffusion. This shared design enables visual and textual information to interact at every layer, allowing the model to infer text priors internally instead of relying on an external OCR module. Compared with prior multi-branch diffusion systems, DualTSR offers a simpler end-to-end formulation with fewer hand-crafted components. Experiments on synthetic Chinese benchmarks and a curated real-world evaluation protocol show that DualTSR achieves strong perceptual quality and text fidelity.
\end{abstract}

\section{Introduction}
\label{sec:intro}

Scene Text Image Super-Resolution (STISR) aims to reconstruct high-quality scene text images from low-resolution inputs. The task is especially challenging because character structures are sensitive to degradation: small distortions, missing strokes, or blending with background clutter can alter the underlying semantics. Unlike general image super-resolution, which primarily enhances textures, STISR must simultaneously guarantee \emph{textual fidelity} and \emph{stylistic realism}. Any structural error directly causes recognition failure, while inconsistencies in font, color, or orientation reduce visual plausibility. These difficulties are further amplified in languages with large character sets such as Chinese.

To reduce the inherent ambiguity of STISR, recent work~\cite{chen2021scene, ma2023text, ma2023benchmark, wang2020scene} incorporates textual priors extracted from a pre-trained OCR model. Knowing the expected text sequence significantly constrains the solution space and often improves reconstruction quality. This OCR-guided paradigm has therefore become dominant, and several recent methods~\cite{noguchi2024scene, zhang2024diffusion} explicitly condition the SR process on predicted text tokens to enforce semantic correctness. However, its reliability is fundamentally limited by the accuracy of the external OCR: erroneous predictions propagate into the SR network, leading to hallucinated strokes, incorrect glyphs, and overall degraded results. To mitigate this dependency, subsequent works introduced stronger low-level priors. MARCONet~\cite{li2023learning} learns a discrete codebook of character structures to provide robust structural guidance, while GlyphSR~\cite{wei2025glyphsr} goes further by generating explicit glyph masks using SAM to supervise fine-grained stroke reconstruction. Although these structural priors alleviate the brittleness of OCR guidance, they require increasingly complex and multi-stage pipelines for prior extraction. This growing reliance on external modules highlights a core limitation of existing approaches and motivates a unified framework that can learn both semantic and structural cues internally without handcrafted or externally produced priors.

To this end, diffusion-based generative models~\cite{songdenoising, lipman2023flow} have recently been adopted for STISR, particularly for complex scripts~\cite{noguchi2024scene, zhang2024diffusion}. DiffTSR~\cite{zhang2024diffusion} reports strong results by modeling text and image with separate modules connected through a multi-modality fusion block. While effective, this multi-stage architecture is cumbersome: it increases system size, complicates training, and limits the depth of cross-modal interaction because textual and visual information are only exchanged at specific fusion points.

To address the reliance on external OCRs and the complexity of multi-module designs, we propose \texttt{DualTSR}, an end-to-end STISR framework built upon a single multimodal transformer. \texttt{DualTSR} unifies super-resolution and text recognition within a dual diffusion objective: a Conditional Flow Matching model generates the high-resolution image, while a discrete diffusion model predicts the character sequence. These two processes operate within a shared architecture, enabling the model to infer its own text priors directly from visual cues and to exploit bidirectional interactions between the evolving text and image representations. This yields a single-model, deeply fused alternative to prior modular pipelines.

Our contributions are threefold:
\begin{itemize}
\item We present \texttt{DualTSR}, a unified STISR framework that jointly learns image super-resolution and text recognition within a single multimodal transformer.
\item We design a dual diffusion objective, continuous (flow matching) for images and discrete diffusion for text, that removes the need for external OCR priors and avoids complex multi-stage architectures.
\item Experiments on synthetic Chinese benchmarks and a curated real-world evaluation protocol show that \texttt{DualTSR} delivers strong perceptual quality and text reconstruction accuracy while preserving a unified end-to-end design.
\end{itemize}

\begin{figure*}[t]
\begin{center}
 \includegraphics[width= \linewidth]{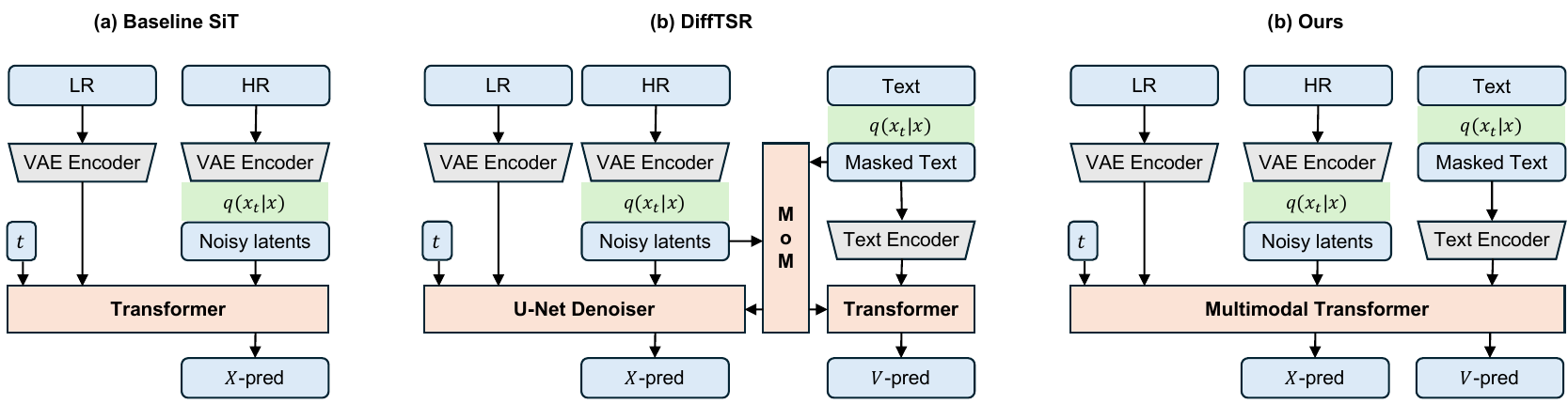}
\end{center}
\caption{\textbf{A comparison of text super-resolution architectures.} Our proposed method unifies image generation and text recognition within a single model.
(a) Baseline: Directly models the conditional distribution of high-resolution images from low-resolution inputs, without a text-based prior.
(b) DiffTSR~\citep{zhang2024diffusion}: Independently models the image and text priors, subsequently fusing their representations using a Multi-modality Module (MoM).
(c) Proposed Method: A unified multimodal transformer jointly optimizes for image generation and text recognition through two distinct but interconnected diffusion processes.}
\vspace{-4mm}
\label{fig:framework}
\end{figure*}

\section{Related Work}
\subsection{Text Image Super-Resolution}
Research in Text Image Super-Resolution (TISR) addresses the unique challenges of simultaneously enhancing image details and improving text legibility. While general-purpose blind super-resolution methods such as BSRGAN~\cite{wu2023practical} and Real-ESRGAN~\cite{wang2021real} restore natural images well, they often fail to preserve the fine-grained structures and appearance cues required for text. This has motivated specialized STISR methods that introduce text-specific priors or auxiliary supervision.

Early and mid-stage STISR work explored complementary cues for restoration. Text Gestalt~\citep{chen2022text} emphasizes stroke-aware reconstruction, TATT~\citep{ma2022text} introduces text attention to improve robustness to spatial deformation, and C3-STISR~\citep{zhao2022c3} combines multiple clues to recover both semantics and structure. A large line of work then adopts OCR-guided priors, where a pre-trained recognizer predicts text that is fed into the SR network to enforce semantic correctness~\citep{wang2020scene, ma2023text, noguchi2024scene, zhang2024diffusion}. These methods can substantially improve recognition-oriented metrics, but their performance remains tied to the quality of the external OCR prior.

More recent methods instead emphasize structural priors. MARCONet~\citep{li2023learning} learns a character-structure codebook, MARCONet++~\citep{li2025enhanced} further strengthens this line with improved character priors, and GlyphSR~\citep{wei2025glyphsr} uses explicit glyph masks generated from an auxiliary pipeline. These approaches reduce dependence on raw OCR predictions, but they also introduce additional stages, modules, or prior-generation procedures. Our method differs from both families: rather than injecting a separately predicted text prior or a handcrafted structural prior, \textbf{DualTSR} learns image restoration and text prediction jointly within one multimodal backbone.

\subsection{Diffusion Models in Super-Resolution}
Diffusion probabilistic models have demonstrated state-of-the-art performance in image synthesis and restoration, owing to their powerful ability to model complex data distributions~\cite{ho2020denoising, lipman2023flow}. This success has naturally led to their application in both natural image super-resolution~\cite{niu2024acdmsr,shang2024resdiff,moser2024diffusion} and TISR ~\cite{zhao2024pean,zhou2024recognition,liu2025textdiff}. Furthermore, research has shown that diffusion models are not only suitable for continuous data like images but can also effectively model discrete data distributions, such as text sequences~\cite{austin2021structured,meng2022concrete,lou2023discrete,sahoo2024simple}.

In the context of TISR, diffusion models have been used in several ways to mitigate unreliable priors and ambiguous degradations. PEAN~\citep{zhao2024pean} refines text-aware features with diffusion-based priors, TextDiff~\citep{liu2023textdiff} uses mask-guided residual diffusion for restoration, and recognition-guided or text-conditional diffusion models~\citep{zhou2024recognition, noguchi2024scene} strengthen the coupling between restoration and text prediction. DiffTSR~\citep{zhang2024diffusion} further separates image generation and text generation into dedicated diffusion branches connected by a multi-modality module, which is particularly effective for Chinese scene text. In contrast, \textbf{DualTSR} uses a single multimodal transformer with two coupled objectives: flow matching for the image branch and discrete diffusion for the text branch. This design keeps the joint modeling ability of diffusion-based STISR while reducing dependence on multiple specialized sub-networks and explicit cross-model fusion modules.


\section{Method}
We introduce \textbf{DualTSR}, an end-to-end multi-modal diffusion framework for scene text image super-resolution. Our approach utilizes a unified backbone to jointly model the distribution of high-resolution images and their corresponding textual content. Given a low-resolution input image $\mathbf{x}^{\text{lr}}$, our objective is to model the conditional distribution $p(\mathbf{x}^{\text{hr}},\mathbf{x}^{\text{txt}}|\mathbf{x}^{\text{lr}})$, enabling the simultaneous generation of a high-resolution image $\mathbf{x}^{\text{hr}}$ and its embedded text $\mathbf{x}^{\text{txt}}$. As illustrated in Figure~\ref{fig:framework}, the DualTSR architecture comprises three core components: (1) a conditional flow matching model for high-resolution image synthesis $p(\mathbf{x}^{\text{hr}}|\mathbf{x}^{\text{lr}},\mathbf{x}^{\text{txt}})$; (2) a discrete diffusion model for optical character recognition (OCR) text generation $p(\mathbf{x}^{\text{txt}}|\mathbf{x}^{\text{lr}},\mathbf{x}^{\text{hr}})$; and (3) a multimodal transformer that unifies these processes to learn the joint distribution $p(\mathbf{x}^{\text{hr}},\mathbf{x}^{\text{txt}}|\mathbf{x}^{\text{lr}})$.

\subsection{Preliminaries}

Our model builds upon two key generative modeling paradigms: conditional flow matching for continuous data like images, and discrete diffusion for textual data.

\subsubsection{Conditional Flow Matching for Image Generation}
Flow matching~\citep{lipman2023flow, tong2024improving} is a recent generative modeling paradigm that formulates sample generation as solving an ordinary differential equation (ODE) defined by a time-dependent velocity field.
At inference, the goal is to generate a high-resolution image $\mathbf{x}_0$ by solving an ODE over a time interval $t \in [0, 1]$. The process starts with a random noise sample $\mathbf{x}_1$ drawn from a standard normal distribution, $p_1(\mathbf{x}_1) = \mathcal{N}(0, I)$. The ODE solver then integrates a learned conditional velocity field, $\mathbf{v}_\theta(\mathbf{x}_t, t, \mathbf{c})$, which guides the transformation from noise to an image. This velocity field is conditioned on external information $\mathbf{c}$ (in our case, the low-resolution image and optionally text labels) and is parameterized by the image head of our multimodal network.

During training, the network $\mathbf{v}_\theta$ is optimized to predict the velocity of a simple linear trajectory between a noise sample $\mathbf{x}_1$ and a target data sample $\mathbf{x}_0$. The path is defined as
\begin{equation}
\mathbf{x}_t = (1-t)\mathbf{x}_0 + t\mathbf{x}_1,
\label{eq:img_forward}
\end{equation}
and its corresponding constant velocity is $\bm{u}_t = \mathbf{x}_1 - \mathbf{x}_0$. The CFM objective minimizes the discrepancy between the predicted velocity and this ground-truth velocity:
\begin{equation}
\mathcal{L}_{\text{CFM}} = \mathbb{E}_{t, q(\mathbf{x}_1), q(\mathbf{x}_0, \mathbf{c})} \left\| \mathbf{v}_\theta(\mathbf{x}_t, t, \mathbf{c}) - \bm{u}_t \right\|^2
\label{eq:cfm_loss}
\end{equation}
where $t$ is sampled uniformly from $[0, 1]$, $q(\mathbf{x}_1)= \mathcal{N}(0, I)$ is the standard normal distribution, and $q(\mathbf{x}_0, \mathbf{c})$ samples from the training dataset.

\subsubsection{Discrete Diffusion for Text Generation}
\label{sec:discrete_diffusion}
For text generation, we require a diffusion process tailored for discrete data, where each token $\mathbf{x}$ belongs to a finite vocabulary $\mathcal{X} = \{1, \dots, N\}$. While some methods apply diffusion in a continuous latent space~\citep{li2022diffusion,chen2022analog,dieleman2022continuous,lovelace2024latent,gulrajani2024likelihood}, this can introduce mapping errors. We instead adopt a diffusion process that operates directly in the discrete token space, a paradigm that has shown strong empirical results \citep{austin2021structured,meng2022concrete,lou2023discrete,sahoo2024simple}.

Our approach is based on a continuous-time Markov chain (CTMC) that defines a forward corruption process. Specifically, we use an absorbing-state diffusion process~\citep{lou2023discrete, sahoo2024simple, shi2024maskdiff}, where tokens in the original text sequence $\mathbf{x}$ are progressively replaced by a special mask token $\mathbf{m}$. The marginal distribution of the corrupted text $\mathbf{x}_t$ at time $t$ is a categorical distribution conditioned on the original text $\mathbf{x}$:
\begin{equation}
    \label{eq:mdlm_xt}
    q(\mathbf{x}_t | \mathbf{x}) = \text{Cat}[\mathbf{x}_t | \alpha_t \mathbf{x} + (1 - \alpha_t) \mathbf{m}],
\end{equation}
where $\text{Cat}(\cdot|\boldsymbol{\pi})$ is the categorical distribution with probabilities $\boldsymbol{\pi}$, and $\alpha_t$ is a noise schedule.

The goal of the reverse process is to learn a denoising distribution $p_\theta(\mathbf{x}\mid \mathbf{x}_t, t)$ that predicts the original clean text $\mathbf{x}$ from its corrupted version $\mathbf{x}_t$. Following recent work \citep{sahoo2024simple, shi2024maskdiff}, we train the model by directly predicting the denoised variate. This leads to a simplified negative variational lower bound (NELBO) objective under the continuous-time limit:
\begin{equation}
\label{eq:mdlm_elbo}
    L_{\text{NELBO}} = \mathbb{E}_{q(\mathbf{x}_t|\mathbf{x})}
    \left[
    \int_{0}^1 \frac{-\alpha_t'}{1-\alpha_t}\log p_\theta(\mathbf{x}\mid \mathbf{x}_t, t)\, dt
    \right].
\end{equation}
In practice, this integral is approximated using Monte-Carlo sampling. For the noise schedule, we follow \citet{sahoo2024simple} and use a simple log-linear schedule where $\alpha_t = 1-t$.

\begin{figure}
  \begin{center}
    \includegraphics[width=0.6\linewidth]{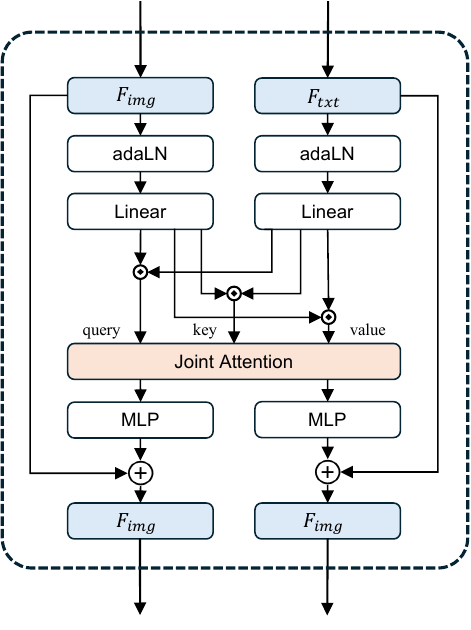}
  \end{center}
  \caption{\textbf{Illustration of the joint attention mechanism.}
Following the MM-DiT block design in SD3~\citep{esser2024scaling}, we incorporate a joint attention module that enables the image latent refinement and text reconstruction processes to operate in parallel.}
  \label{fig:joint_attention}
  \vspace{-4mm}
\end{figure}

\subsection{Multimodal Transformer}

To enable effective interaction between image and text modalities, we adopt a multimodal transformer architecture inspired by the MM-DiT block design from SD3~\citep{esser2024scaling}. Unlike prior DiffTSR approaches~\citep{zhang2024diffusion}, which rely on separate modality-specific models connected through an auxiliary communication module, our framework employs a unified transformer backbone. This unified design allows the model to dynamically attend to different modalities depending on the input, thereby supporting joint optimization of both the super-resolution task (LR-to-HR) and the recognition task (LR-to-text).

At the heart of this design is a joint attention mechanism~\citep{esser2024scaling}. As illustrated in Figure~\ref{fig:joint_attention}, latent tokens from the image and text diffusion processes are first processed independently. During the attention step, tokens from both modalities are concatenated to construct shared query, key, and value representations, which are passed through a single self-attention block. The resulting outputs are then split and routed back into their respective modality-specific streams. This mechanism achieves deep multimodal fusion, enabling visual features to directly guide text reconstruction and textual cues to inform image super-resolution at every layer of the network.
\subsection{Training}
\label{sec:training}
To enable synchronized image and text generation at inference time, we optimize the continuous image process and the discrete text process jointly. Given a training triple $(\mathbf{x}^{\text{hr}}, \mathbf{x}^{\text{lr}}, \mathbf{x}^{\text{txt}})$, the HR image is encoded by a variational autoencoder~\citep{9578911} into the latent space used by the flow-matching branch, while the text sequence is embedded into discrete tokens. For simplicity, we keep the notation $\mathbf{x}^{\text{hr}}$ and $\mathbf{x}^{\text{lr}}$ for the corresponding latent variables in the equations below.

Let $\mathbf{v}_\theta$ denote the image-velocity head and $p_\theta$ denote the text-denoising head of the shared multimodal transformer. We sample
\[
\mathbf{x}^{\text{hr}}_{t} = (1-t)\mathbf{x}^{\text{hr}} + t\bm{\epsilon}, \qquad \bm{\epsilon}\sim\mathcal{N}(0,I),
\]
and corrupt the text with the absorbing process in Equation~\ref{eq:mdlm_xt} to obtain $\mathbf{x}^{\text{txt}}_t$. The corresponding image target velocity is $\mathbf{u}_t=\bm{\epsilon}-\mathbf{x}^{\text{hr}}$.

We first optimize modality-specific conditional objectives. For image generation, the model predicts the HR velocity conditioned on the LR image and the clean text:
\begin{align}
\mathcal{L}_{\text{IMG}} = \mathbb{E}_{t,q} \left\| \mathbf{v}_{\theta}\left( \mathbf{x}^{\text{hr}}_{t}, t, \mathbf{c}^{\text{img}} \right) - \mathbf{u}_t \right\|_2^2,
\end{align}
where $\mathbf{c}^{\text{img}}=\{\mathbf{x}^{\text{lr}}, \mathbf{x}^{\text{txt}}\}$. For text prediction, the model reconstructs the clean text from corrupted tokens conditioned on the LR image and the clean HR image:
\begin{align}
\mathcal{L}_{\text{TXT}} = \mathbb{E}_{q^{(\text{txt})}} \left[
- \frac{1}{K} \sum_{i=1}^{K} \log p_\theta\left(\mathbf{x}^{\text{txt}} \mid \mathbf{x}^{\text{txt}}_{t_i}, t_i, \mathbf{c}^{\text{txt}} \right)
\right],
\end{align}
where $\mathbf{c}^{\text{txt}}=\{\mathbf{x}^{\text{lr}}, \mathbf{x}^{\text{hr}}\}$. Following~\citet{Li_2025_CVPR}, we approximate the continuous-time objective with antithetic sampling~\citep{NEURIPS2021_b578f2a5} over $K$ timesteps $t_i$ uniformly covering $(\delta,1]$, where $\delta$ is a small constant for numerical stability.

To strengthen joint modeling, we further corrupt both modalities with the same timestep and train the model to recover them simultaneously:
\begin{align}
\mathcal{L}_{\text{Joint}} = \mathbb{E}_{t,q^{(\text{hr,lr,txt})}} \Big[
&\left\| \mathbf{v}_{\theta}\left( \mathbf{x}^{\text{hr}}_{t}, t, \tilde{\mathbf{c}}^{\text{img}} \right) - \mathbf{u}_t \right\|_2^2 \nonumber\\
&- \log p_\theta\left(\mathbf{x}^{\text{txt}} \mid \mathbf{x}^{\text{txt}}_{t}, t, \tilde{\mathbf{c}}^{\text{txt}} \right)
\Big],
\end{align}
where $\tilde{\mathbf{c}}^{\text{img}}=\{\mathbf{x}^{\text{lr}}, \mathbf{x}^{\text{txt}}_t\}$ and $\tilde{\mathbf{c}}^{\text{txt}}=\{\mathbf{x}^{\text{lr}}, \mathbf{x}^{\text{hr}}_t\}$. The overall training objective is
\begin{equation}
\mathcal{L}_{\text{Overall}} = \mathcal{L}_{\text{IMG}} + \mathcal{L}_{\text{TXT}} + \mathcal{L}_{\text{Joint}}.
\end{equation}

\subsection{Model-Guided Training via Classifier-Free Guidance}

To improve sample quality and controllability at inference time, we incorporate model guidance (MG) directly into training, following \citet{tang2025diffusion}. We apply this idea to the image-flow target in Equation~\ref{eq:cfm_loss}. Let $\mathbf{v}_{\theta}^{\text{ema}}\left( \mathbf{x}_{t}, t, \bm{c} \right)$ be the conditioned velocity predicted by the EMA teacher model and $\mathbf{v}_{\theta}^{\text{ema}}\left( \mathbf{x}_{t}, t, \varnothing \right)$ be the corresponding unconditional prediction. The flow-matching objective is then reformulated as
\begin{equation}
    \mathcal{L}_{\text{CFM-MG}} = \mathbb{E}_{t, q(\mathbf{x}_0), q(\mathbf{x}_1, \mathbf{c})} \left\| \mathbf{v}_\theta(\mathbf{x}_{t}, t, \mathbf{c}) - \bm{u}'_t \right\|^2
\end{equation}
where the rectified target $\bm{u}'_t$ is defined as:
\begin{equation}
    \bm{u}'_t(\mathbf{x}_{t}, t,\bm{c}) = \bm{u}_t + w \cdot \left( \text{sg}\left(\mathbf{v}^{\text{ema}}_\theta(\mathbf{x}_{t}, t,\bm{c})\right) - \mathbf{v}^{\text{ema}}_\theta(\mathbf{x}_{t}, t, \varnothing) \right),
    \label{eq:mg_target}
\end{equation}
where $w$ denotes the guidance scale factor and $\text{sg}(\cdot)$ is the stop-gradient operator. Using the EMA teacher stabilizes the target prediction throughout training. During optimization, the condition vector $\bm{c}$ is randomly replaced with $\varnothing$ with probability $\psi$ so that the network learns both conditional and unconditional predictions.

Applying this rectified target to the image-only and joint image losses gives
\begin{equation}
\begin{aligned}
    &\mathcal{L}_{\text{IMG-MG}} =   \mathbb{E}_{t,q^{(\text{img})}} \\
    & \left\| \mathbf{v}_{\theta}\left( \mathbf{x}^{\text{hr}}_{t}, t, \mathbf{c}^{\text{img}} \right) - \bm{u}'_t(\mathbf{x}^{\text{hr}}_{t}, t,\mathbf{c}^{\text{img}}) \right\|_2^2, \\
    &\mathcal{L}_{\text{Joint-MG}} = 
    \mathbb{E}_{t,q^{(\text{img,txt})}} \Big[
    \left\| \mathbf{v}_{\theta}\left( \mathbf{x}^{\text{hr}}_{t}, t, \tilde{\mathbf{c}}^{\text{img}} \right) - \bm{u}'_t(\mathbf{x}^{\text{hr}}_{t}, t,\tilde{\mathbf{c}}^{\text{img}}) \right\|_2^2 \\
    &\hspace{32mm} - \log p_\theta\left(\mathbf{x}^{\text{txt}} \mid \mathbf{x}^{\text{txt}}_{t}, t, \tilde{\mathbf{c}}^{\text{txt}} \right)\Big].
\end{aligned}
\end{equation}
Our final training objective is then defined as:
\begin{equation}
\mathcal{L}_{\text{Overall}} = \mathcal{L}_{\text{IMG-MG}} + \mathcal{L}_{\text{TXT}} + \mathcal{L}_{\text{Joint-MG}}.
\end{equation}

\subsection{Joint inference}

At inference time, given a low-resolution image $\mathbf{x}^{\text{lr}}$, we initialize the image branch from Gaussian noise $\mathbf{x}^{\text{hr}}_1$ and the text branch from a fully masked sequence $\mathbf{x}^{\text{txt}}_1$. The shared transformer then updates both modalities synchronously. For a step from $t$ to $s$ ($0<s<t<1$), the model predicts
\begin{align}
\hat{\mathbf{u}}_t &= \mathbf{v}_{\theta}\left(\mathbf{x}^{\text{hr}}_{t}, t, \{\mathbf{x}^{\text{lr}}, \mathbf{x}^{\text{txt}}_t\}\right), \\
\mathbf{p}_t &= p_\theta\left(\cdot \mid \mathbf{x}^{\text{txt}}_t, t, \{\mathbf{x}^{\text{lr}}, \mathbf{x}^{\text{hr}}_t\}\right).
\end{align}
The image latent is updated with an Euler step,
\begin{equation}
\mathbf{x}^{\text{hr}}_{s} = \mathbf{x}^{\text{hr}}_{t} - (t-s)\hat{\mathbf{u}}_t,
\end{equation}
while the text latent is updated by the reverse absorbing-state transition parameterized by $\mathbf{p}_t$, with only masked positions being resampled and already generated tokens kept fixed. After obtaining the final image latent $\mathbf{x}^{\text{hr}}_0$, we decode it back to RGB space using the VAE decoder. Detailed pseudocode for training and sampling is provided in Appendix~\ref{app:pseudo_code}.

\begin{table*}[ht]
    \centering
    \caption{\textbf{Quantitative comparison on the synthetic CTR-TSR}. Best in \textbf{bold}.}
    \label{tab:ctr_results}
    \resizebox{1.0\hsize}{!}{
    \begin{tabular}{c|ccccc|ccccc}
    \toprule
         \multirow{2}{*}{\bf Method} & \multicolumn{5}{c|}{$\times$\textbf{2}} & \multicolumn{5}{c}{$\times$\textbf{4}} \\ \cline{2-11}
        &\bf PSNR$\uparrow$& \bf LPIPS$\downarrow$& \bf FID$\downarrow$& \bf ACC$\uparrow$& \bf NED$\uparrow$ &\bf PSNR$\uparrow$& \bf LPIPS$\downarrow$& \bf FID$\downarrow$& \bf ACC$\uparrow$& \bf NED$\uparrow$\\
        \midrule
        ESRGAN~\citep{wang2018ESRGAN} & 24.09 & 0.3046 & 15.06 & 67.08\% & 83.79\% & 22.18 & 0.3986 & 18.25 & 43.69\% & 62.15\% \\
        MSRResNet~\citep{9022144} & 28.03 & 0.3030 & 30.47 & 68.94\% & 85.37\% & 24.52 & 0.4029 & 50.60 & 49.04\% & 67.96\% \\
        SwinIR~\citep{liang2021swinir} & 28.39 & 0.2983 & 31.54 & 70.01\% & 86.10\% & 24.73 & 0.3957 & 50.89 & 50.09\% & 68.93\% \\
        SRFormer~\citep{zhou2023srformer} & \textbf{28.89} & 0.2829 & 26.93 & 70.93\% & 86.91\% & \textbf{25.05} & 0.3801 & 46.23 & 51.83\% & 70.70\% \\ \hline
        MARCONet~\citep{li2023learning} & 23.14 & 0.4518 & 88.06 & 57.93\% & 76.43\% & 25.05 & 0.3801 & 46.23 & 51.83\% & 70.70\% \\
        DiffTSR~\citep{zhang2024diffusion} & 23.21 & 0.3304 & 18.57 & 63.51\% & 80.99\% & 20.62 & 0.3952 & 22.24 & 44.87\% & 63.20\% \\
        MARCONet++~\citep{li2025enhanced} & 24.02 & 0.4281 & 64.27 & 59.34\% & 77.49\% & 22.06 & 0.5072 & 86.18 & 39.92\% & 58.30\% \\
        \hline
        \cc{\textbf{DualTSR (Ours)}} & \cc{22.43} & \cc{\textbf{0.2682}} & \cc{\textbf{8.73}} & \cc{\textbf{73.23\%}} & \cc{\textbf{88.50\%}} & \cc{20.54} & \cc{\textbf{0.3292}} & \cc{\textbf{16.42}} & \cc{\textbf{57.65\%}} & \cc{\textbf{76.64\%}} \\
        \bottomrule
    \end{tabular}}
\end{table*}

\begin{table*}[ht]
    \centering
    \caption{\textbf{Quantitative comparison on RealCE}. Best in \textbf{bold}.}
    \label{tab:realce_results}
    \resizebox{1.0\hsize}{!}{
    \begin{tabular}{c|ccccc|ccccc}
    \toprule
         \multirow{2}{*}{\bf Method} & \multicolumn{5}{c|}{$\times \textbf{2}$} & \multicolumn{5}{c}{$\times \textbf{4}$} \\ \cline{2-11}
        &\bf PSNR$\uparrow$& \bf LPIPS$\downarrow$& \bf FID$\downarrow$& \bf ACC$\uparrow$& \bf NED$\uparrow$ &\bf PSNR$\uparrow$& \bf LPIPS$\downarrow$& \bf FID$\downarrow$& \bf ACC$\uparrow$& \bf NED$\uparrow$\\
        \midrule
        ESRGAN~\citep{wang2018ESRGAN} & 20.08 & 0.3833 & 103.50 & 58.80\% & 86.30\% & 19.97 & 0.3721 & 87.18 & 58.50\%& 87.65\% \\
        MSRResNet~\citep{9022144} & 19.25 & \textbf{0.3033} & 43.82 & 62.80\% & 89.08\% & 19.28 & 0.3590 & 60.81 & 58.00\%& 86.87\% \\
        SwinIR~\citep{liang2021swinir} & 20.24 & 0.3810 & 87.98 & 59.30\% & 86.44\% & 20.23 & 0.3271 & 56.77 & 62.10\%& 87.55\% \\
        SRFormer~\citep{zhou2023srformer} & \textbf{20.85} & 0.3081 & 50.81 & 62.90\% & 89.47\% & 20.29 & 0.3732 & 83.47 & 57.50\%& 85.79\% \\
        \hline
        MARCONet~\citep{li2023learning} & 18.93 & 0.4144 & 87.54 & 60.58\% & 87.11\% & \textbf{20.30} & 0.3732 & 83.47 & 57.50\%& 85.79\% \\
        DiffTSR~\citep{zhang2024diffusion} & 18.95 & 0.3179 & 38.13 & 62.60\% & 88.13\% & 18.09 & 0.3382 & 41.13 & 58.00\%& 84.44\% \\
        MARCONet++~\citep{li2025enhanced} & 19.38 & 0.3839 & 75.47 & 61.20\% & 87.47\% & 19.48 & 0.3953 & 81.72 & 59.40\%& 87.76\% \\
        \hline
        \cc{\textbf{DualTSR (Ours)}} & \cc{18.94} & \cc{0.3133} & \cc{\textbf{35.95}} & \cc{\textbf{63.80\%}} & \cc{\textbf{89.97\%}} & \cc{18.85} & \cc{\textbf{0.3277}} & \cc{\textbf{40.78}} & \cc{\textbf{62.20\%}} & \cc{\textbf{88.49\%}} \\
        \bottomrule
    \end{tabular}}
    \vspace{-2mm}
\end{table*}

\section{Experiments}
\label{sec:experiments}

\subsection{Experimental Setup}

\textbf{Datasets.}
We evaluate the effectiveness of our method on two benchmarks.
First, we construct a Chinese scene text super-resolution dataset, denoted as \textbf{CTR-TSR}.
Following DiffTSR~\citep{zhang2024diffusion}, we create training and testing splits from CTR~\citep{yu2021benchmarking}; detailed construction procedures (filtering, canonical HR resizing, and blind degradation) are provided in Appendix~\ref{app:ctr-tsr-construction}.
Because the exact DiffTSR split is unavailable, we re-implemented the preprocessing pipeline per their description to ensure comparability.
Second, we evaluate on the \textbf{RealCE} benchmark~\citep{ma2023benchmark}, which contains real-world Chinese text images. In practice, adapting the full benchmark to paired line-level STISR is difficult because many samples exhibit partial annotations, inaccurate localization, or severe LR--HR misalignment. We therefore report paired metrics on a curated subset of 300 samples with clean and well-aligned LR--HR pairs. These RealCE results should be interpreted under this explicit protocol rather than as a claim of full-benchmark superiority.

\noindent\textbf{Evaluation Metrics.}
For both CTR-TSR and RealCE, we report PSNR (↑), LPIPS (↓), FID (↓), and two text fidelity metrics: ACC (↑) and NED (↑)~\citep{zhang2024diffusion}.
ACC is the exact-match recognition accuracy (word-level); NED measures character level similarity by normalizing the Levenshtein distance~\citep{zhang2024diffusion}.
To avoid evaluator drift, we use a fixed pretrained OCR recognizer (TransOCR from CTR~\citep{yu2021benchmarking}) for all methods unless otherwise noted; computation details for the text metric are in Appendix~\ref{app:text-metrics}.

\noindent\textbf{Degradation and Resolutions.}
Unless specified, HR images are resized to $128\times512$ (text-line layout preserved). LR images are synthesized by a blind, stochastic pipeline (blur + noise + compression + downsampling, with random ordering/magnitudes) adapted from BSRGAN~\cite{zhang2021designing} and Real-ESRGAN~\cite{wang2021real}; full recipe and randomization ranges are summarized in Appendix~\ref{app:ctr-tsr-construction}.

\noindent\textbf{Training Configuration.}
Our model is implemented in PyTorch and trained on 4$\times$NVIDIA A100 (40 GB) GPUs.
We use AdamW~\citep{loshchilov2018decoupled} with learning rate $1\times 10^{-4}$, cosine decay, and weight decay $0.05$.
For CTR-TSR, we train for 700k iterations with batch size 32 (global).
By default, we adopt $K=8$ discrete-text timesteps for antithetic sampling in the text diffusion loss and a 4-step ODE sampler for joint inference unless otherwise stated. Unless noted otherwise, we set the guidance scale factor $w$ in Equation~\ref{eq:mg_target} to 1.0, which provides the best overall trade-off in Table~\ref{tab:cfg_ablation}. More experiment details are provided in Appendix~\ref{app:exp_detail}. Code, dataset construction scripts, and pretrained checkpoints will be released.
\noindent\textbf{Baselines.} We compare with general SR methods (ESRGAN~\citep{wang2018ESRGAN}, MSRResNet~\citep{9022144}, SwinIR~\citep{liang2021swinir}, SRFormer~\citep{zhou2023srformer}) and text-specific approaches (MARCONet~\citep{li2023learning}, MARCONet++~\citep{li2025enhanced}, DiffTSR~\citep{zhang2024diffusion}).
For MARCONet/MARCONet++, we use official checkpoints due to their tailored synthetic pipelines; DiffTSR is also evaluated with the official checkpoint on our test splits. Since these text-specific baselines are not retrained under a fully matched degradation pipeline, the comparison should be read as a reference comparison under a unified evaluation protocol rather than a perfectly controlled retraining study.

\subsection{Main Results}

\noindent\textbf{Quantitative.}
Tables~\ref{tab:ctr_results} and~\ref{tab:realce_results} report performance on the synthetic CTR-TSR benchmark and the real-world RealCE benchmark~\citep{ma2023benchmark}. On CTR-TSR, generic SR models (ESRGAN~\citep{wang2018ESRGAN}, MSRResNet~\citep{9022144}, SwinIR~\citep{liang2021swinir}, SRFormer~\citep{zhou2023srformer}) show strong PSNR but achieve poor LPIPS/FID and low ACC/NED at both $\times2$ and $\times4$, confirming that high-fidelity image reconstruction alone is insufficient for recovering structured text.
MARCONet~\citep{li2023learning} and MARCONet++~\citep{li2025enhanced} introduce text-aware supervision, yet their compositing-based training often disrupts scene consistency and limits perceptual performance. DiffTSR improves FID through its dual diffusion branches, but the separation of text and image pathways still leads to occasional semantic errors. On this synthetic benchmark, DualTSR achieves the strongest FID/LPIPS/ACC/NED values across both $\times2$ and $\times4$ settings, although PSNR remains lower than several reconstruction-oriented baselines. This pattern suggests that the unified multimodal transformer favors perceptual realism and text fidelity over pure pixel-wise similarity.

On the curated RealCE subset, the real-world degradations amplify the weaknesses of prior approaches: generic SR models preserve coarse textures but fail to reconstruct valid glyphs, while MARCONet~\citep{li2023learning}, MARCONet++~\citep{li2025enhanced}, and DiffTSR show limited robustness beyond their synthetic training assumptions. Under this protocol, DualTSR achieves the best ACC and NED at both scales together with competitive perceptual quality. Because prior works often evaluate different filtered subsets or unreleased samples on RealCE, we interpret these results as protocol-specific evidence of transfer to real imagery rather than a definitive ranking on the full benchmark.


\begin{figure*}[!ht]
  \begin{center}
    \includegraphics[width=\linewidth]{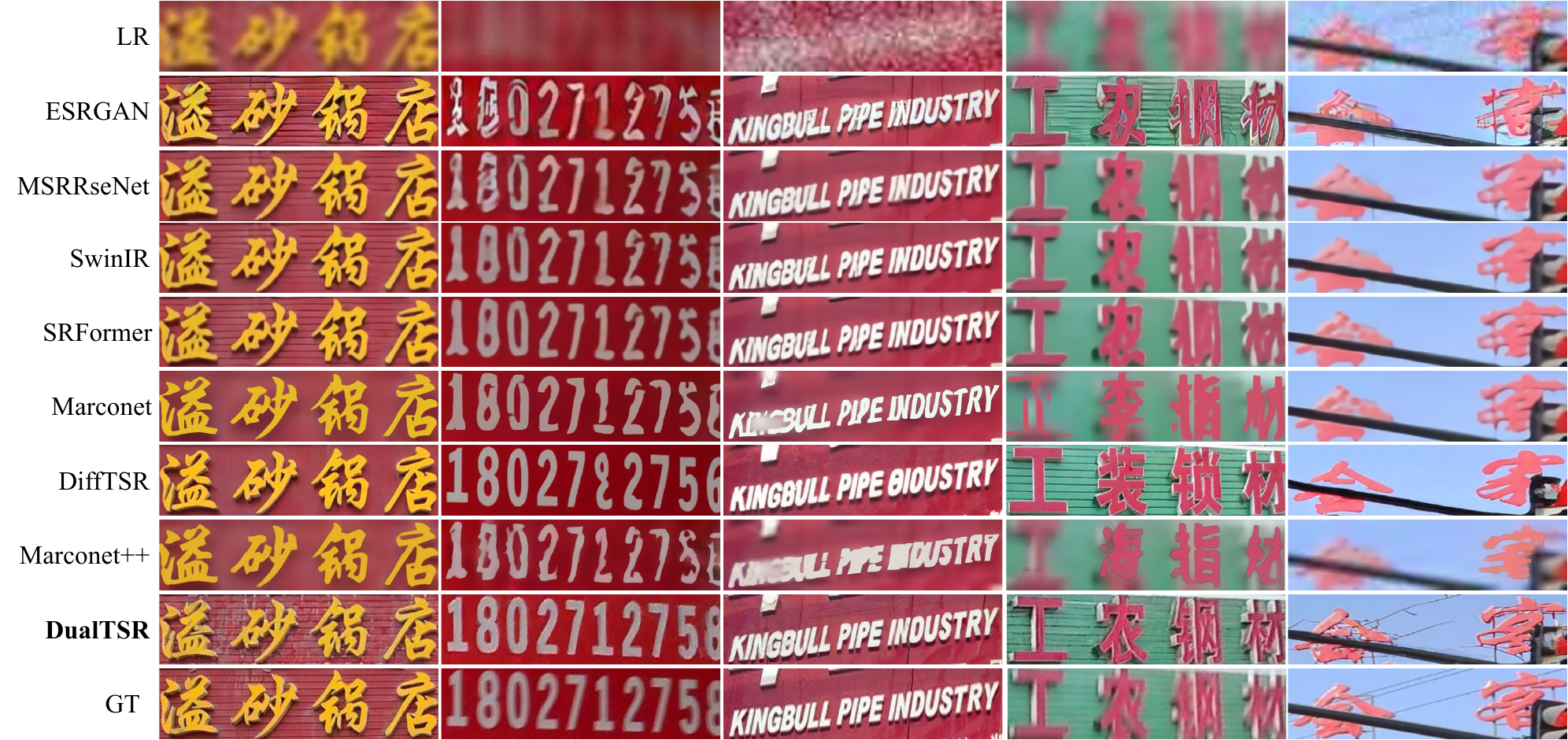}
  \end{center}
  \caption{\textbf{Qualitative comparison for the synthetic dataset CTR-TSR-Test with different methods on $\times$4 scale.}
  The comparison methods include ESRGAN~\citep{wang2018ESRGAN}, MSRResNet~\citep{9022144}, SwinIR~\citep{liang2021swinir}, SRFormer~\citep{zhou2023srformer}, MARCONet~\citep{li2023learning}, MARCONet++~\citep{li2025enhanced}, DiffTSR~\citep{zhang2024diffusion} and our method.}
  \label{fig:ctr_test_X4}
\end{figure*}

\begin{figure*}[!ht]
  \begin{center}
    \includegraphics[width=\linewidth]{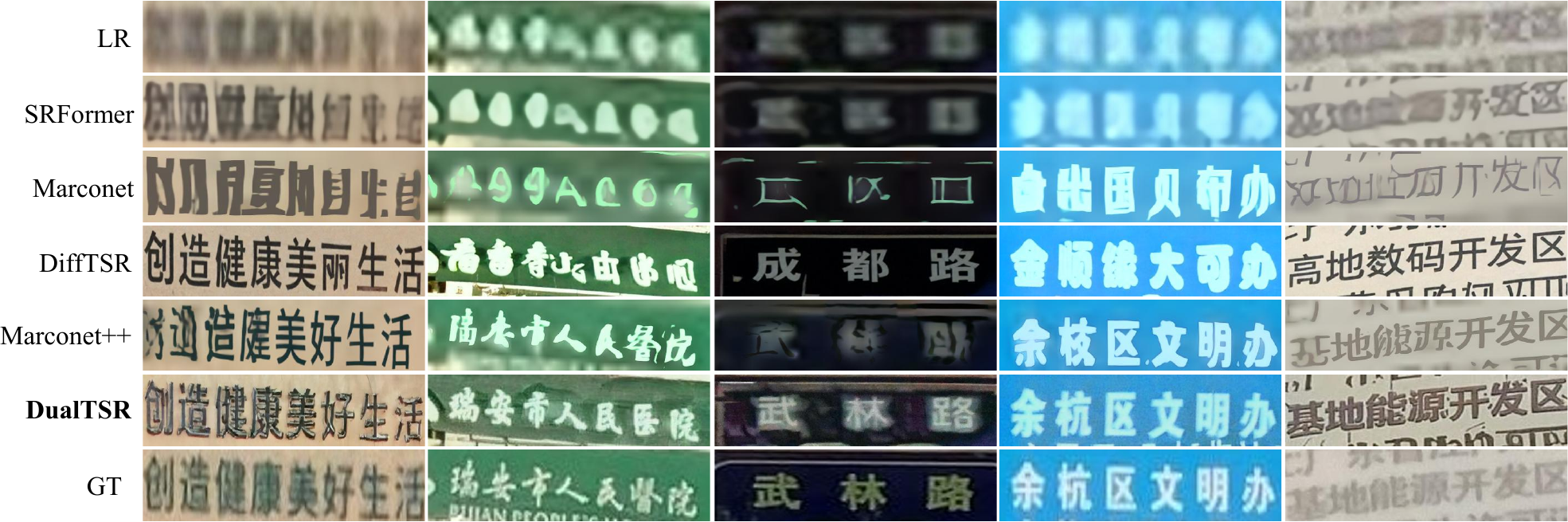}
  \end{center}
  \caption{\textbf{Qualitative comparison for RealCE with different methods on $\times$4 scale.}
  The comparison methods include ESRGAN~\citep{wang2018ESRGAN}, MSRResNet~\citep{9022144}, SwinIR~\citep{liang2021swinir}, SRFormer~\citep{zhou2023srformer}, MARCONet~\citep{li2023learning}, MARCONet++~\citep{li2025enhanced}, DiffTSR~\citep{zhang2024diffusion} and our method.}
  \label{fig:realce_X4}
\end{figure*}

\noindent\textbf{Qualitative.}
Figure~\ref{fig:ctr_test_X4} and Figure~\ref{fig:realce_X4} provide $\times4$ visual comparisons on the CTR-TSR and RealCE datasets.
Generic SR models tend to oversmooth strokes or introduce structural distortions, reflecting their lack of text-awareness. Text-oriented baselines better preserve character shapes, but MARCONet~\citep{li2023learning} and MARCONet++~\citep{li2025enhanced} often generate glyphs that appear stylistically inconsistent or artificially pasted onto the background, despite occasionally producing correct character identities. DiffTSR yields sharper characters, yet its dual-branch design can lead to semantically incorrect predictions. In contrast, DualTSR more often recovers the correct textual content while preserving local appearance cues and keeping the text integrated with the surrounding scene. Failure cases still occur under severe corruption, especially when color or font cues are largely missing, in which case the model may prioritize legibility over exact style preservation. We provide more visual examples and additional $\times2$ qualitative results in Appendix~\ref{app:more_qualit}.

\noindent\textbf{Evaluation scope.}
Our empirical claims are intentionally limited to the settings reported here. The paper evaluates Chinese STISR on CTR-TSR and on a curated RealCE protocol, and several text-specific baselines are compared through official checkpoints rather than matched retraining. Accordingly, we treat the main conclusion as evidence that a unified multimodal backbone can achieve strong text fidelity and perceptual quality without external OCR priors, rather than as a universal ranking over all STISR benchmarks and protocols.

\subsection{Ablation Study}

\paragraph{Loss design ablation study.}
Table~\ref{tab:abl_components} presents an ablation on our loss components. Starting with the baseline {(a) $\mathcal{L}_{\text{Joint-MG}}$}, the model already achieves strong results by jointly corrupting image and text with synchronized timesteps, yielding a balanced performance across both image quality (FID 13.73, LPIPS 0.3804) and text reconstruction (ACC 49.85\%, NED 68.28\%).
Adding the discrete text diffusion loss {(b) $\mathcal{L}_{\text{Joint-MG}} + \mathcal{L}_{\text{TXT}}$} further improves the text metrics substantially (ACC +2.87, NED +3.89) while slightly reducing FID and LPIPS, indicating that explicit supervision on the text pathway stabilizes the shared Transformer and benefits both modalities.

Finally, incorporating the full image–text training objective {(c) $\mathcal{L}_{\text{Joint-MG}} + \mathcal{L}_{\text{TXT}} + \mathcal{L}_{\text{IMG-MG}}$} yields the best overall performance, improving FID from 13.39 to 9.92 and LPIPS from 0.3782 to 0.3550, while also boosting text accuracy to 53.71\%. These gains show that the image MG loss complements the discrete text diffusion: image–level gradient signals regularize the latent Transformer and ensure that text reconstruction aligns with visually plausible high-resolution structures.

Overall, the ablation confirms the \emph{Joint-MG} mechanism is essential for coupling the two modalities, while the additional image/text losses strengthen modality-specific reconstruction. The best results are achieved when all losses operate together, demonstrating that synchronized corruption paired with both continuous (image) and discrete (text) objectives leads to the most coherent multimodal generation.

\begin{table}[t]
    \centering
    \caption{\textbf{Loss ablation study.} All models are trained for 300k iterations with a batch size of 128.}
    \label{tab:abl_components}
    \resizebox{\linewidth}{!}{
    \begin{tabular}{lccccc}
        \toprule
        Setting & FID$\downarrow$ & PSNR$\uparrow$ & LPIPS$\downarrow$ & ACC$\uparrow$ & NED$\uparrow$ \\
        \midrule
        (a) $\mathcal{L}_{\text{Joint-MG}}$ & 13.73&19.82&0.3804&49.85\%&68.28\%\\
        (b) $\mathcal{L}_{\text{Joint-MG}} + \mathcal{L}_{\text{TXT}}$ & 13.39&19.79&0.3782&52.72\%& 72.17\%\\
        {(c) $\mathcal{L}_{\text{Joint-MG}} + \mathcal{L}_{\text{TXT}} + \mathcal{L}_{\text{IMG-MG}}$ }  & 9.92&20.12&0.3550&53.71\%&73.60\%\\
        \bottomrule
    \end{tabular}
    }
\end{table}

\begin{figure}[t]
  \begin{center}
    \includegraphics[width=\linewidth]{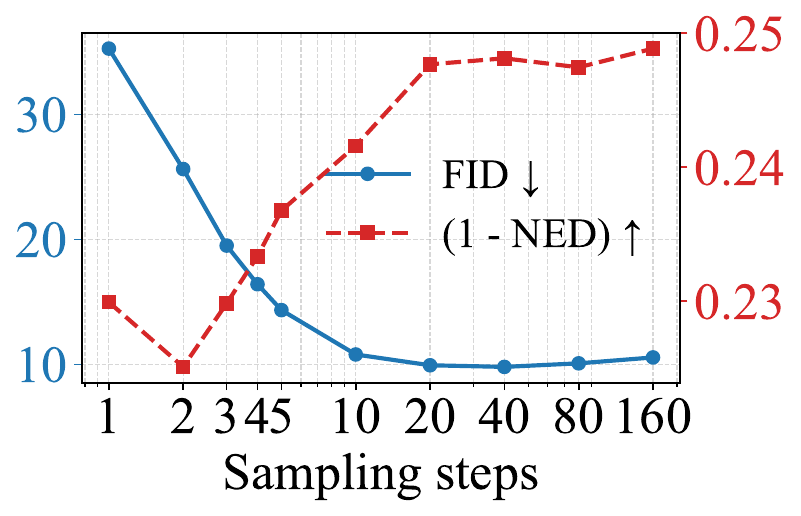}
  \end{center}
\vspace{-4mm}
  \caption{\textbf{Trade-off between perceptual quality and text fidelity across sampling steps.} 
FID improves with more sampling steps, reaching its best value around \(\sim\!40\), while NED worsens as text strokes become over-smoothed. 
We select \(4\) steps as a balanced operating point.}
\vspace{-4mm}
  \label{fig:trade_off_smapling_step}
\end{figure}
\begin{figure}[!ht]
  \begin{center}
    \includegraphics[width=1\linewidth]{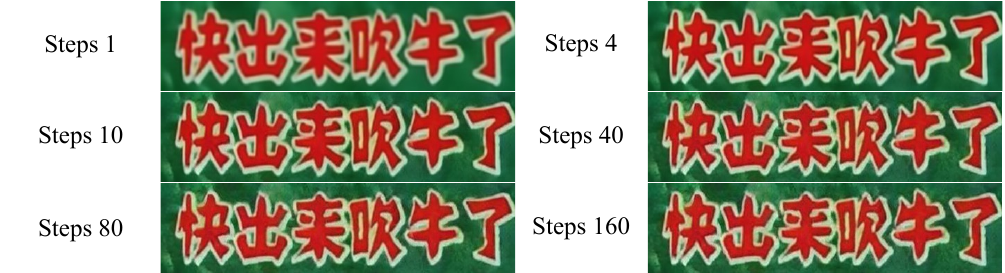}
  \end{center}
\vspace{-4mm}
  \caption{\textbf{Trade-off between perceptual quality and text fidelity across sampling steps.}}
\vspace{-2mm}
  \label{fig:step_visual}
\end{figure}

\paragraph{Effect of sampling steps.}
Figure~\ref{fig:trade_off_smapling_step} illustrates the trade-off between perceptual realism and text fidelity as we vary the {total number of sampling steps} during inference. Increasing the number of steps generally improves FID, with the best perceptual quality achieved around 40 steps, reflecting more effective refinement of textures and global coherence.
In contrast, NED peaks at only 2 steps and degrades steadily as additional refinement is applied, indicating diminished character legibility. We hypothesize that with very few sampling steps, the model focuses on reconstructing coarse structures and high-contrast edges, which are crucial cues for recognizing fine-grained glyph shapes. As the total steps increase, the sampler devotes more iterations to enhancing textural realism and smoothing artifacts (see Fig.~\ref{fig:step_visual}), but this refinement can inadvertently erode subtle, high-frequency stroke patterns—particularly problematic for dense scripts like Chinese where small details distinguish similar characters. Balancing these two factors, we adopt 4 sampling steps as our default setting, which provides a favorable compromise between natural image appearance and accurate text reconstruction.

\begin{table}[ht]
\centering
\caption{Effect of CFG scale on FID and NED. Increasing CFG worsens FID and eventually harms NED.}
\resizebox{\linewidth}{!}{
\begin{tabular}{c|ccccc}
\toprule
\textbf{CFG} & 1 & 1.45 & 2 & 4 & 6 \\
\midrule
\textbf{FID$\downarrow$} & 16.10 & 19.59 & 24.19 & 35.35 & 39.49 \\
\textbf{NED$\uparrow$} & 76.66\% & 76.71\% & 75.95\% & 73.09\% & 72.20\% \\
\bottomrule
\end{tabular}}
\label{tab:cfg_ablation}
\end{table}

\paragraph{Effect of CFG scale.}
Table~\ref{tab:cfg_ablation} shows the impact of classifier-free guidance (CFG) on our model. In contrast to typical image synthesis, where stronger guidance often improves perceptual sharpness~\cite{ho2021classifierfree}, we find that increasing CFG consistently worsens FID, while NED remains nearly unchanged at 1.45 and then declines as guidance becomes stronger. Larger CFG values push the model toward overly confident predictions that deviate from the true data distribution, leading to unnatural textures and distorted character structures. We therefore use a CFG scale of 1.0, which offers the best overall trade-off. We hypothesize that strong guidance amplifies high-frequency hallucinations during sampling, which harms both visual naturalness and the fine-grained stroke patterns essential for accurate text reconstruction.

\section{Conclusion}
\label{sec:conclusion}

We presented \texttt{DualTSR}, a unified framework for text image super-resolution that integrates super-resolution and text recognition within a single multimodal transformer. By jointly modeling continuous image generation and discrete text prediction, our method removes the reliance on external OCR priors and avoids the multi-stage, multi-model design of prior approaches such as DiffTSR. Experiments on CTR-TSR and on our curated RealCE evaluation protocol show that \texttt{DualTSR} provides strong perceptual quality and text fidelity within a single end-to-end model. At the same time, our current study is limited to Chinese benchmarks and checkpoint-based comparisons for some baselines; broader benchmark coverage and fully matched retraining remain important future work. We believe this unified perspective provides a promising direction for robust and scalable scene text restoration.

{
    \small
    \bibliographystyle{ieeenat_fullname}
    \bibliography{main}
}

\clearpage
\newpage
\appendix
{
\onecolumn
\centering
\Large
\textbf{\thetitle}\\
\vspace{1.0em}
}
\section{Appendix}

\subsection{Architectural Comparison}

Beyond quantitative and qualitative performance, our method also differs from prior diffusion-based STISR systems at the architectural level. DiffTSR~\citep{zhang2024diffusion} adopts a multi-stage pipeline composed of separate image and text diffusion branches plus an additional fusion module. Such a design offers strong modeling flexibility, but it also introduces more moving parts during training and evaluation.

In contrast, our framework unifies text recognition and image super-resolution within a single multimodal transformer and optimizes both branches end to end. Table~\ref{tab:complexity_comparison} therefore summarizes structural differences in the training framework rather than measured runtime or memory costs. The key point is that DualTSR reduces the number of independently trained components while retaining joint image-text modeling.

\begin{table*}[h]
\centering
\caption{Comparison of training framework. Our framework consolidates text recognition and super-resolution into a single, end-to-end module, whereas DiffTSR~\citep{zhang2024diffusion} requires a multi-stage pipeline with multiple components and auxiliary models.}
\label{tab:complexity_comparison}
\begin{tabular}{lcc}
\toprule
\textbf{Feature} & \textbf{DiffTSR} & \textbf{Ours} \\
\midrule
\textbf{Training Stages} & 3-stage training & End-to-end training \\
\textbf{Core Components} & 3 (IDM, TDM, MoM)  & 1 (Multimodal Transformer) \\
\textbf{Auxiliary Models} & Pre-trained VAE + text recognizer  & Pre-trained VAE \\
\bottomrule
\end{tabular}
\end{table*}

\subsection{CTR-TSR dataset construction}
\label{app:ctr-tsr-construction}

We build the \textbf{CTR-TSR} dataset from the CTR~\citep{yu2021benchmarking} corpus following the preprocessing and synthesis procedure described in DiffTSR~\citep{zhang2024diffusion}. In brief, the construction pipeline is:

\begin{enumerate}
  \item \textbf{Source images.} Start from the scene part of CTR dataset~\citep{yu2021benchmarking} as the pool of high-quality HR text line images.
  \item \textbf{Filtering and selection.} Retain only HR images that satisfy all of the following criteria:
    \begin{itemize}
      \item resolution (longer side) $\geq 64$ pixels;
      \item width-to-height ratio $> 2$ (to focus on text line images);
      \item length of text annotation $\leq 24$ characters.
    \end{itemize}
    After filtering and resizing (next step), this yields 64139 HR images which we denote {CTR-TSR-Train}.
    
  \item \textbf{HR normalization.} Resize each selected HR image to a canonical HR size of $128\times512$ for training (preserving the overall layout of text lines).
  \item \textbf{LR synthesis (degradation).} We generate LR images using a blind, high-complexity degradation pipeline inspired by BSRGAN~\citep{zhang2021designing} and Real-ESRGAN~\citep{wang2021real}. During training, one of the two degradation strategies is randomly selected with equal probability ($p=0.5$) for each sample, and the corresponding LR image is synthesized online.
  \item \textbf{Synthetic test set.} For the synthetic evaluation set ({CTR-TSR-Test}) we select images from the scene CTR test split, apply the same filtering/resizing and the same degradation pipeline as above. The resulting CTR-TSR-Test contains 8690 LR–HR pairs.
\end{enumerate}

All synthesized LR images used for training and synthetic testing are generated with the same stochastic degradation pipeline so that training and evaluation are comparable across methods.

\subsection{Additional experiment details}
\label{app:exp_detail}
The quantitative results in Tables~\ref{tab:ctr_results} and~\ref{tab:realce_results} use CFG $=1.0$ and 4 sampling steps. The qualitative results in Figures~\ref{fig:ctr_test_X4} and~\ref{fig:realce_X4} use CFG $=1.0$ and 160 sampling steps.

\subsection{Training and Sampling Pseudocode}
\label{app:pseudo_code}
Here, we provide pseudocode for the DualTSR training process in Algorithm~\ref{alg:training} and the inference process in Algorithm~\ref{alg:inference}.

\renewcommand{\lstlistingname}{Algorithm}

\begin{figure}[t]
\begin{lstlisting}[style=pytorchstyle, label={alg:training}, caption={PyTorch-style pseudocode for the proposed DualTSR training strategy. We utilize an EMA teacher model to provide guidance targets for the flow matching objective.}]
def train_step(model, model_ema, optimizer, batch, w=1.0, psi=0.1):
    x_hr, x_lr, text = batch  # Unpack data batch
    
    # 1. Image-Only Guided Loss (Flow Matching + Model Guidance)
    t = torch.rand(x_hr.shape[0])
    x_1 = torch.randn_like(x_hr)           # Sample Noise
    x_t = (1 - t) * x_hr + t * x_1         # Linear Interpolation
    
    # Calculate Guidance Target using EMA Teacher
    with torch.no_grad():
        out_cond = model_ema(x_t, t, cond={'img': x_lr, 'txt': text})
        out_unc  = model_ema(x_t, t, cond={'img': x_lr, 'txt': None})
        
        # Target = Base_Velocity + w * (Cond_Pred - Uncond_Pred)
        u_target = (x_1 - x_hr) + w * (out_cond - out_unc)

    # Student Prediction (with CFG dropout)
    cond = {'img': x_lr, 'txt': text} if rand() > psi else {'img': x_lr, 'txt': None}
    u_pred = model(x_t, t, cond)
    loss_img = F.mse_loss(u_pred, u_target)

    # 2. Text-Only Loss (Masked Generative Training)
    t_txt = sample_timesteps()
    text_masked = mask_tokens(text, ratio=1-t_txt)
    
    logits_txt = model(text_masked, t_txt, cond={'img': x_hr}, mode='text')
    loss_txt = F.cross_entropy(logits_txt, text) / t_txt

    # 3. Joint Guided Loss (Simultaneous Generation)
    t_j = torch.rand(x_hr.shape[0])
    x_t_j = (1 - t_j) * x_hr + t_j * x_1
    txt_t_j = mask_tokens(text, ratio=1-t_j)

    with torch.no_grad():
        # Joint guidance targets from EMA
        c_j = {'txt_in': txt_t_j, 'lr': x_lr}
        u_j_c = model_ema(x_t_j, t_j, c_j)
        u_j_u = model_ema(x_t_j, t_j, {'txt_in': None, 'lr': x_lr})
        u_target_j = (x_1 - x_hr) + w * (u_j_c - u_j_u)

    c_stud = c_j if rand() > psi else {'txt_in': None, 'lr': x_lr}
    u_pred_j, logits_j = model(x_t_j, t_j, c_stud, mode='joint')
    loss_joint = F.mse_loss(u_pred_j, u_target_j) + F.cross_entropy(logits_j, text)

    # Optimization Step
    loss = loss_img + loss_txt.mean() + loss_joint
    loss.backward()
    optimizer.step()
    update_ema(model, model_ema)
\end{lstlisting}
\label{alg:pytorch_code}
\end{figure}

\begin{figure}[t]
\begin{lstlisting}[style=pytorchstyle, label={alg:inference}, caption={PyTorch-style inference algorithm. The image is generated via ODE flow integration, while the text is generated via iterative unmasking (discrete diffusion) synchronized with the image generation steps.}]
def joint_inference(model, x_lr, steps=50, seq_len=25):
    # 1. Initialization
    bs = x_lr.shape[0]
    x = torch.randn_like(x_lr)                     # Image Latent (t=1)
    text = torch.full((bs, seq_len), MASK_TOKEN)   # Text Latent (All Masked)

    # Time schedule from 1.0 down to 0.0
    timesteps = torch.linspace(1.0, 0.0, steps + 1)
    dt = 1.0 / steps

    # 2. Generation Loop
    for k in range(steps):
        t = timesteps[k]
        s = timesteps[k+1]  # Next timestep

        # Predict Velocity and Text Logits
        u_pred, logits_txt = model(x, t, text, cond=x_lr)

        # --- A. Image Update (Euler ODE Step) ---
        # Reverse the flow: subtract velocity to go from Noise -> Data
        x = x - dt * u_pred

        # --- B. Text Update (Reverse Discrete Diffusion) ---
        # Determine probability to unmask at this step
        alpha_t, alpha_s = 1 - t, 1 - s
        prob_unmask = (alpha_s - alpha_t) / (1.0 - alpha_t + 1e-8)

        # Sample candidate tokens from model prediction
        probs = F.softmax(logits_txt, dim=-1)
        candidates = torch.multinomial(probs.view(-1, probs.size(-1)), 1).view(bs, seq_len)

        # Update Logic:
        # If token is [MASK] AND random < prob_unmask -> Update with candidate
        # Else -> Keep current state (either already generated or still [MASK])
        is_masked = (text == MASK_TOKEN)
        should_sample = torch.rand_like(text.float()) < prob_unmask
        
        update_mask = is_masked & should_sample
        text[update_mask] = candidates[update_mask]

    return x, text
\end{lstlisting}
\label{alg:inference_code}
\end{figure}

\subsection{Reproduction Details for Different Methods}
For MSRResNet~\citep{9022144} and ESRGAN~\citep{wang2018ESRGAN}, we use the BasicSR\footnote{https://github.com/XPixelGroup/BasicSR/tree/master} codebase to train the models on our CTR-TSR-Train dataset with the default settings. For SRFormer~\citep{zhou2023srformer}, we use the official code and training settings from the original paper. For DiffTSR~\citep{zhang2024diffusion}, we use the official checkpoint for direct inference on our test splits. For MARCONet~\citep{li2023learning} and MARCONet++~\citep{li2025enhanced}, we use the publicly available checkpoints because these models rely on carefully designed data-synthesis pipelines and are intended to generalize to real-world scenarios. Since DiffTSR, MARCONet, and MARCONet++ are not retrained under an identical degradation pipeline, these comparisons should be interpreted as reference comparisons under a common evaluation protocol rather than as fully controlled retraining results. We observe that MARCONet++ requires detectable text in the LR image; when the model fails to detect text, inference raises an error. In such cases, we use the LR image directly as the model output. This issue is especially common in real-world samples whose text can barely be recognized.

\subsection{Protocol Note on RealCE}
RealCE was originally released for real-world text image super-resolution, but paired metric evaluation is difficult when adapting it to line-level Chinese STISR because some samples contain partial annotations, inaccurate localization, or severe LR--HR misalignment. For this reason, we report paired metrics on a curated subset of 300 clean LR--HR pairs. This protocol improves measurement stability and reproducibility, but the resulting numbers should be understood as protocol-specific rather than as a full-benchmark ranking.

\subsection{Text-Related Metrics}
\label{app:text-metrics}

\paragraph{ACC (word accuracy / recognition accuracy).}
ACC is the exact-match recognition accuracy: the fraction of test samples for which the predicted text sequence equals the ground-truth transcription. In all experiments in this paper, we use the pretrained TransOCR recognition model from~\cite{yu2021benchmarking} (as in DiffTSR) to obtain predictions used in ACC calculation. Higher ACC indicates better preservation of textual content in the restored image.

\paragraph{NED (Normalized Edit Distance).}
NED quantifies character-level similarity by normalizing the Levenshtein (edit) distance between the predicted and ground-truth strings. For a single sample with ground truth $g$ and prediction $p$ we compute:
\[
\mathrm{NED}(p,g) = 1 - \frac{\mathrm{ED}(p,g)}{\max(|g|,\,|p|)},
\]
where $\mathrm{ED}(\cdot,\cdot)$ is the Levenshtein distance and $|g|$, and $|p|$ is the length of the ground truth and prediction. We report the dataset mean NED; larger values indicate better (closer) recognition. In CTR-TSR evaluations ACC and NED are computed using the pretrained TransOCR recognizer, as in DiffTSR.

\subsection{More Qualitative examples}
\label{app:more_qualit}
\subsubsection{On synthetic CTR-TSR}
\begin{figure}[H]
  \begin{center}
    \includegraphics[width=\linewidth]{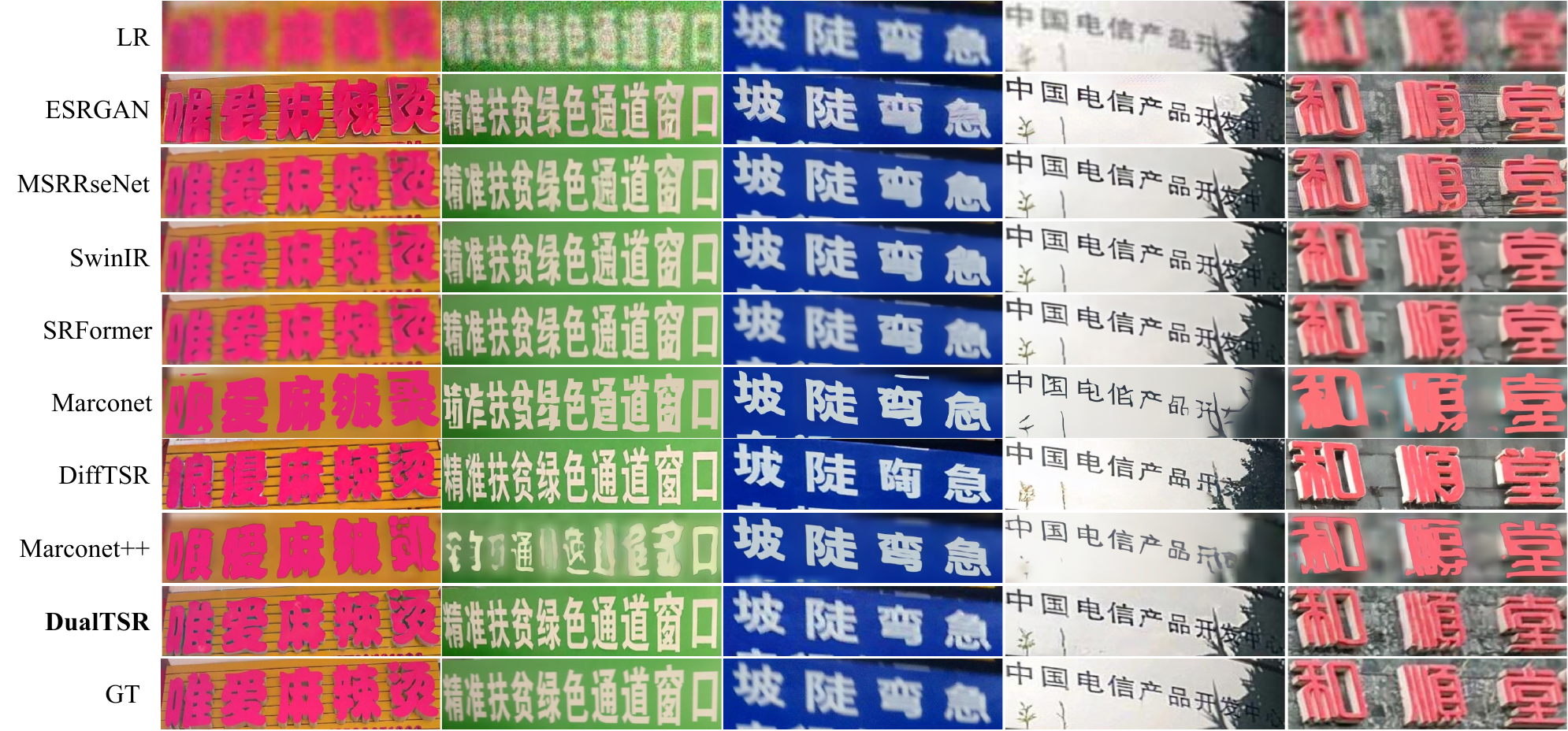}
  \end{center}
  \caption{\textbf{Visual comparison for the synthetic dataset CTR-TSR-Test with different methods on $\times2$ scale.}
  The comparison methods include ESRGAN~\citep{wang2018ESRGAN}, MSRResNet~\citep{9022144}, SwinIR~\citep{liang2021swinir}, SRFormer~\citep{zhou2023srformer}, MARCONet~\citep{li2023learning}, MARCONet++~\citep{li2025enhanced}, DiffTSR~\citep{zhang2024diffusion} and our method.}\label{fig:CTR_x2_1}
\end{figure}

\begin{figure}[H]
  \begin{center}
    \includegraphics[width=\linewidth]{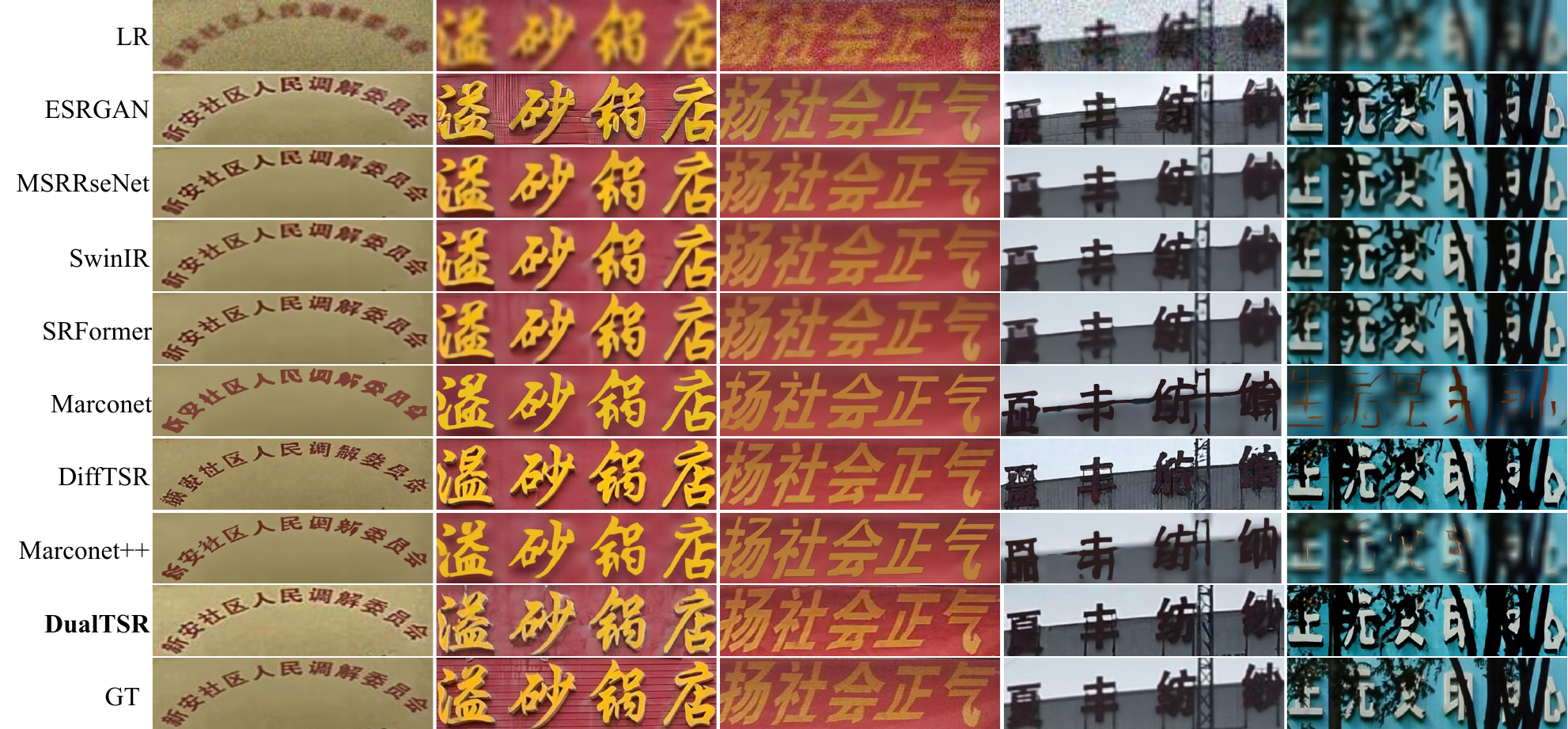}
  \end{center}
  \caption{\textbf{Visual comparison for the synthetic dataset CTR-TSR-Test with different methods on $\times2$ scale.}
  The comparison methods include ESRGAN~\citep{wang2018ESRGAN}, MSRResNet~\citep{9022144}, SwinIR~\citep{liang2021swinir}, SRFormer~\citep{zhou2023srformer}, MARCONet~\citep{li2023learning}, MARCONet++~\citep{li2025enhanced}, DiffTSR~\citep{zhang2024diffusion} and our method.}
  \label{fig:CTR_x2_2}
\end{figure}

\begin{figure}[H]
  \begin{center}
    \includegraphics[width=\linewidth]{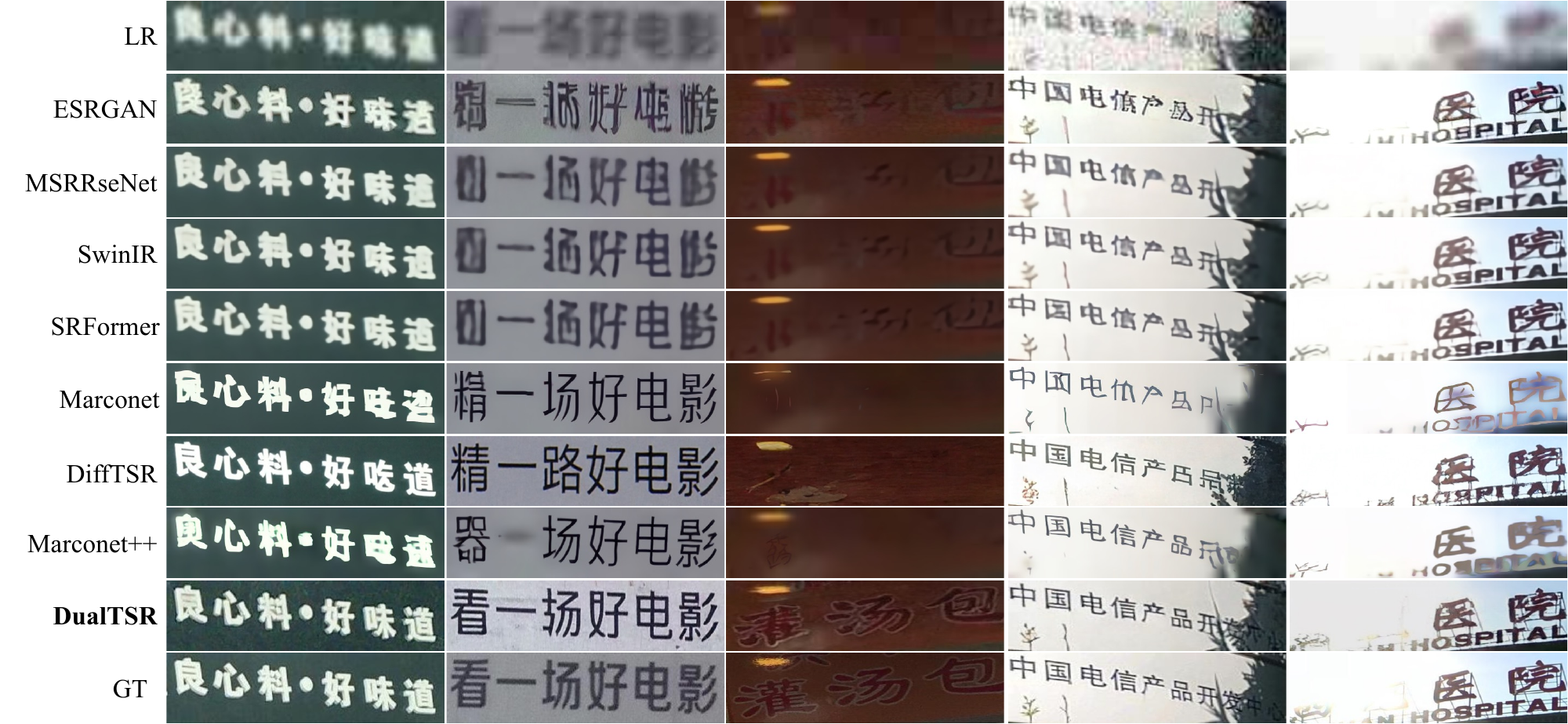}
  \end{center}
  \caption{\textbf{Visual comparison for the synthetic dataset CTR-TSR-Test with different methods on $\times4$ scale.}
  The comparison methods include ESRGAN~\citep{wang2018ESRGAN}, MSRResNet~\citep{9022144}, SwinIR~\citep{liang2021swinir}, SRFormer~\citep{zhou2023srformer}, MARCONet~\citep{li2023learning}, MARCONet++~\citep{li2025enhanced}, DiffTSR~\citep{zhang2024diffusion} and our method.}
  \label{fig:CTR_x4_2}
\end{figure}

For the synthetic CTR-TSR test set, we provide additional $\times2$ results in Figures~\ref{fig:CTR_x2_1} and~\ref{fig:CTR_x2_2}, and additional $\times4$ results in Figure~\ref{fig:CTR_x4_2}.


\subsubsection{On real-world dataset RealCE}
\begin{figure}[H]
  \begin{center}
    \includegraphics[width=\linewidth]{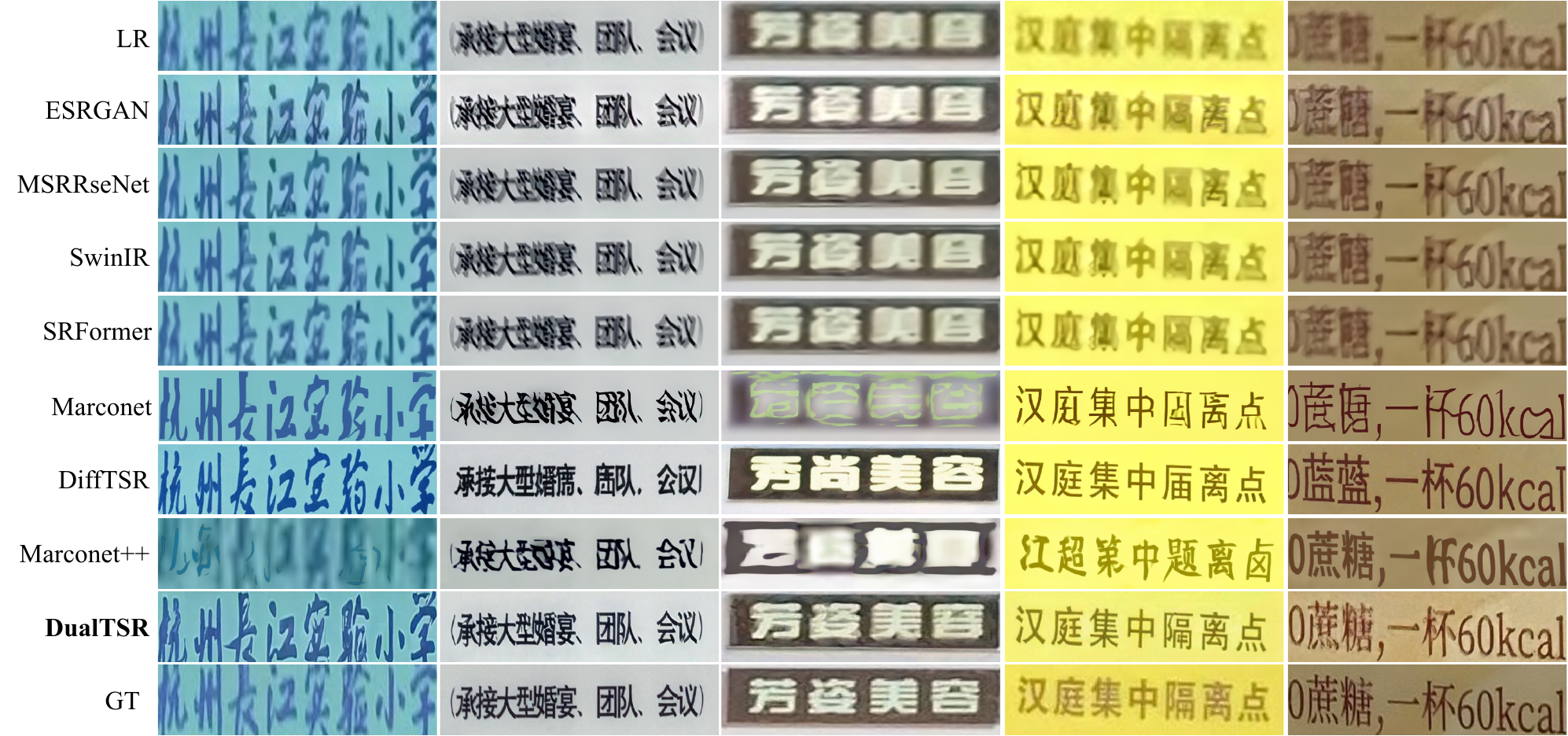}
  \end{center}
  \caption{\textbf{Visual comparison for the real-world dataset RealCE with different methods on $\times2$ scale.}
  The comparison methods include ESRGAN~\citep{wang2018ESRGAN}, MSRResNet~\citep{9022144}, SwinIR~\citep{liang2021swinir}, SRFormer~\citep{zhou2023srformer}, MARCONet~\citep{li2023learning}, MARCONet++~\citep{li2025enhanced}, DiffTSR~\citep{zhang2024diffusion} and our method.}\label{fig:realce_ours_x2_1}
\end{figure}

\begin{figure}[H]
  \begin{center}
    \includegraphics[width=\linewidth]{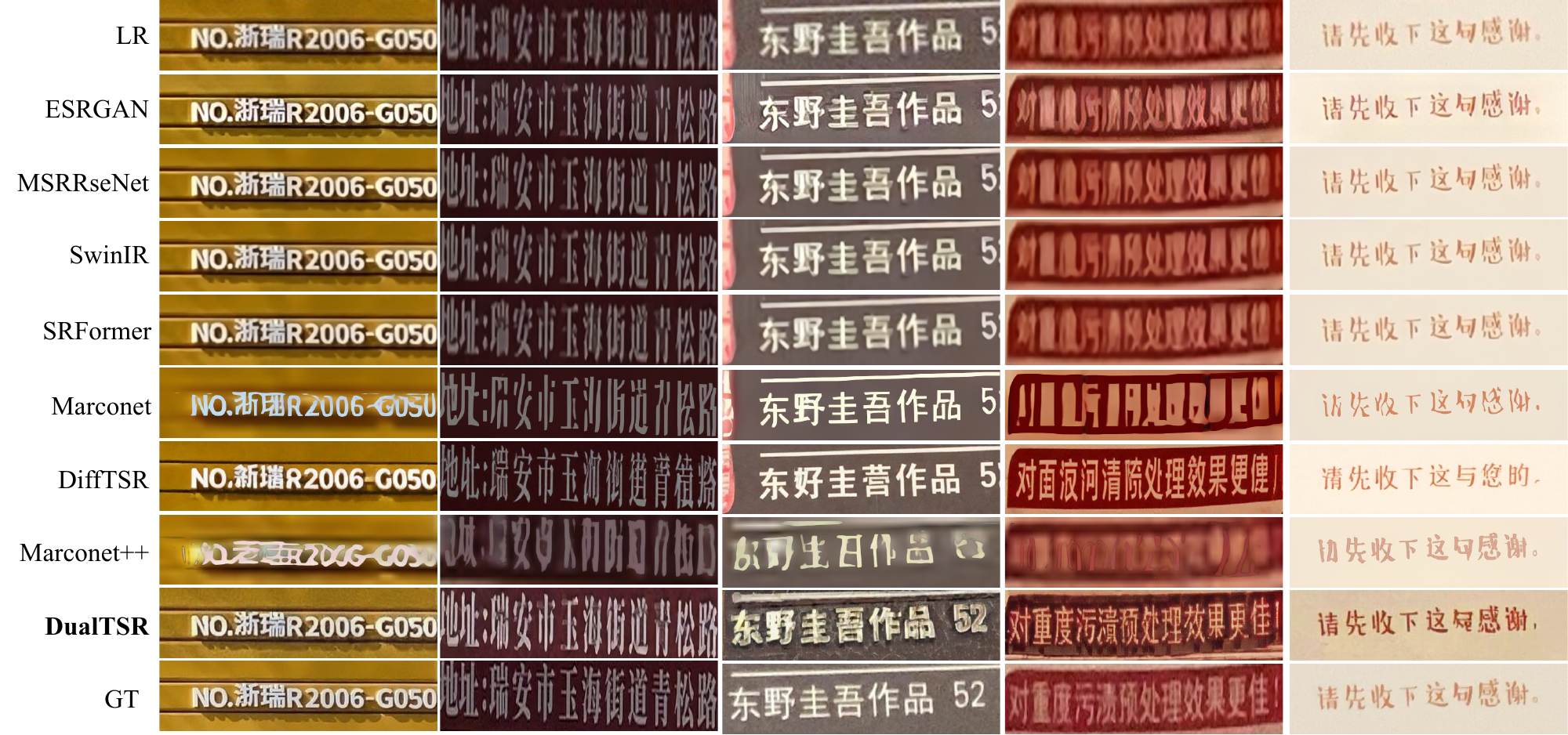}
  \end{center}
  \caption{\textbf{Visual comparison for the real-world dataset RealCE with different methods on $\times2$ scale.}
  The comparison methods include ESRGAN~\citep{wang2018ESRGAN}, MSRResNet~\citep{9022144}, SwinIR~\citep{liang2021swinir}, SRFormer~\citep{zhou2023srformer}, MARCONet~\citep{li2023learning}, MARCONet++~\citep{li2025enhanced}, DiffTSR~\citep{zhang2024diffusion} and our method.}\label{fig:realce_ours_x2_2}
\end{figure}

\begin{figure}[H]
  \begin{center}
    \includegraphics[width=\linewidth]{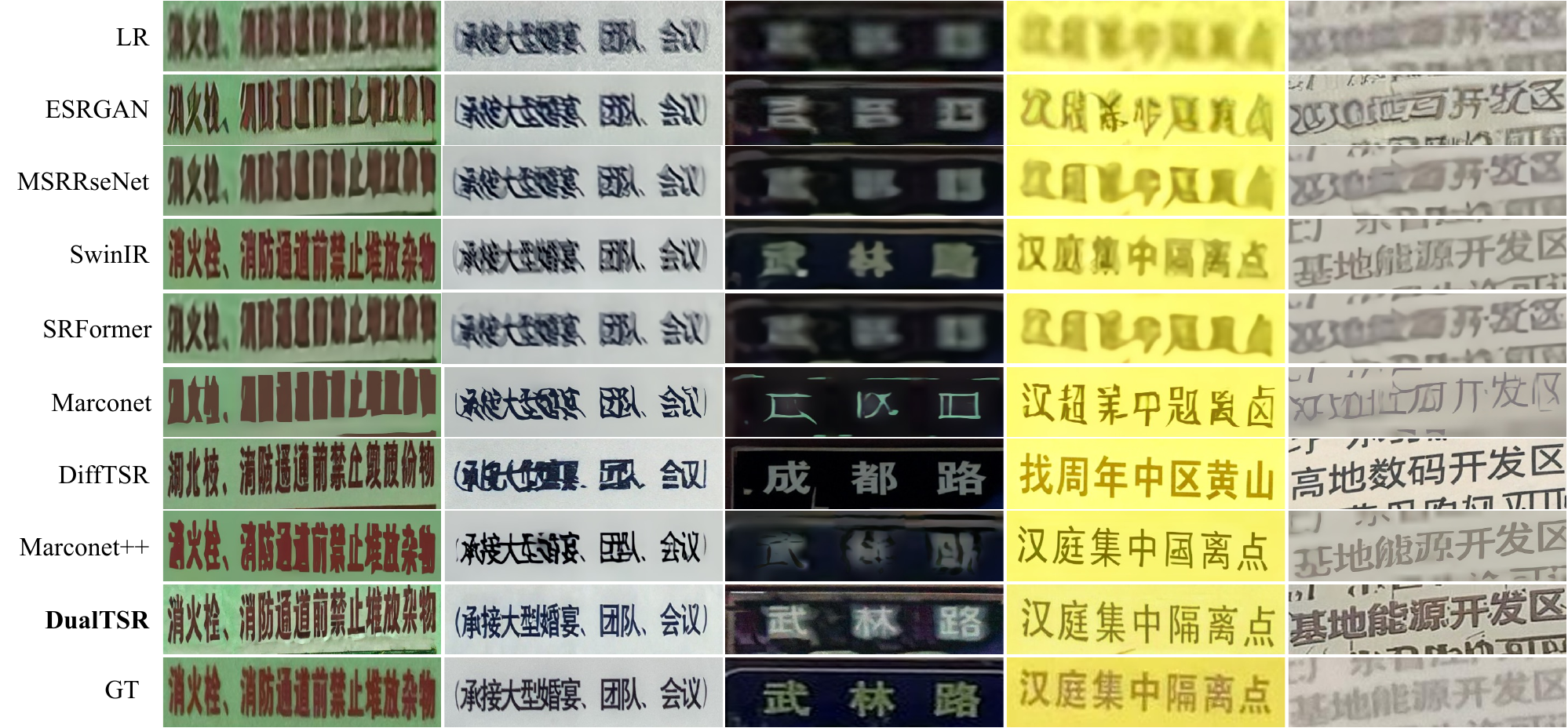}
  \end{center}
  \caption{\textbf{Visual comparison for the real-world dataset RealCE with different methods on $\times4$ scale.}
  The comparison methods include ESRGAN~\citep{wang2018ESRGAN}, MSRResNet~\citep{9022144}, SwinIR~\citep{liang2021swinir}, SRFormer~\citep{zhou2023srformer}, MARCONet~\citep{li2023learning}, MARCONet++~\citep{li2025enhanced}, DiffTSR~\citep{zhang2024diffusion} and our method.}\label{fig:realce_ours_x4_2}
\end{figure}

\begin{figure}[H]
  \begin{center}
    \includegraphics[width=\linewidth]{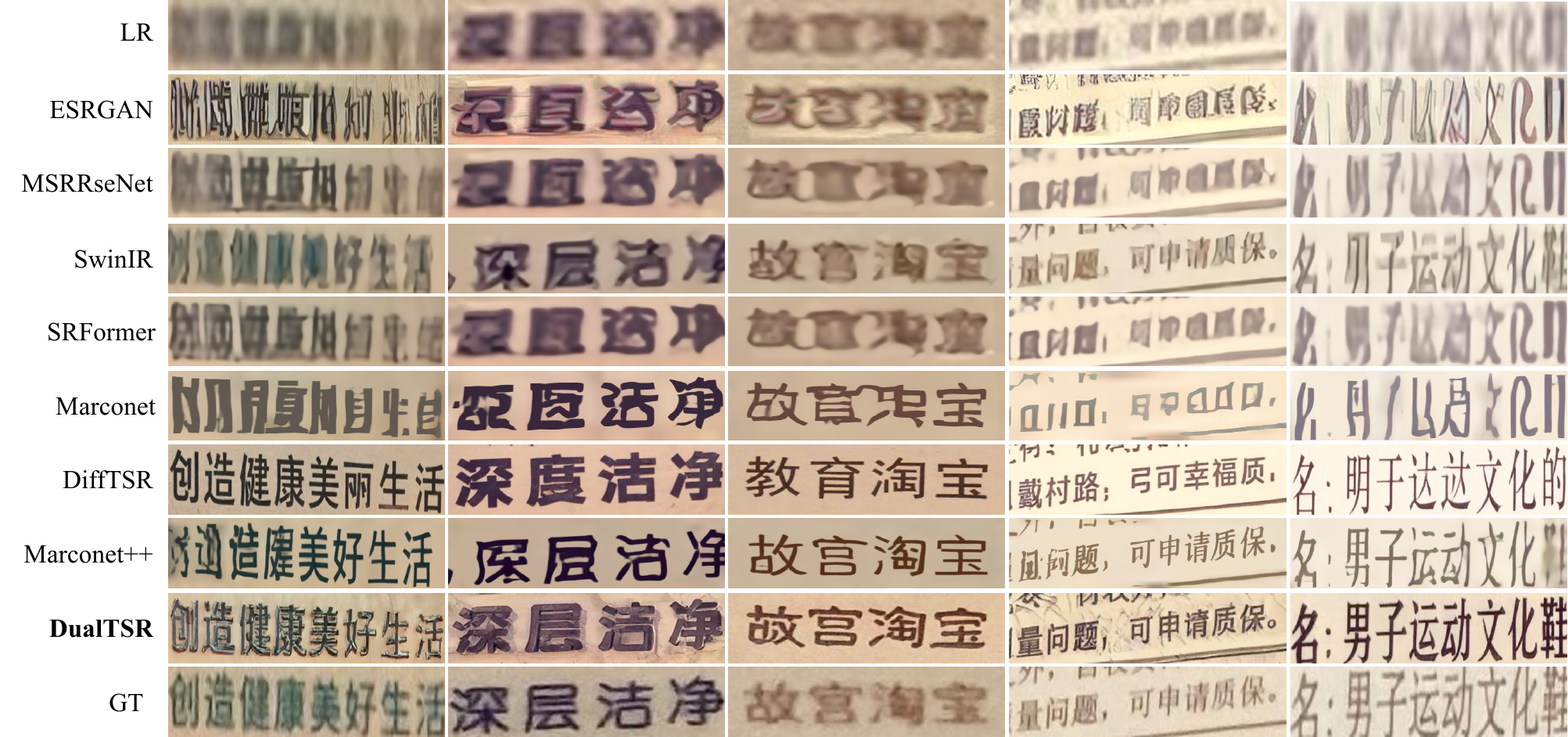}
  \end{center}
  \caption{\textbf{Visual comparison for the real-world dataset RealCE with different methods on $\times4$ scale.}
  The comparison methods include ESRGAN~\citep{wang2018ESRGAN}, MSRResNet~\citep{9022144}, SwinIR~\citep{liang2021swinir}, SRFormer~\citep{zhou2023srformer}, MARCONet~\citep{li2023learning}, MARCONet++~\citep{li2025enhanced}, DiffTSR~\citep{zhang2024diffusion} and our method.}\label{fig:realce_ours_x4_3}
\end{figure}

For the real-world RealCE benchmark, we provide additional $\times2$ results in Figures~\ref{fig:realce_ours_x2_1} and~\ref{fig:realce_ours_x2_2}, and additional $\times4$ results in Figures~\ref{fig:realce_ours_x4_2} and~\ref{fig:realce_ours_x4_3}.


\end{document}